\documentclass[sigplan,screen]{acmart}
\renewcommand\footnotetextcopyrightpermission[1]{} 
\makeatletter
\renewcommand\@formatdoi[1]{\ignorespaces}
\makeatother
\setcopyright{none}
\renewcommand\footnotetextcopyrightpermission[1]{}
\usepackage{multirow}
\usepackage[
  separate-uncertainty = true,
  multi-part-units = repeat
]{siunitx}
\AtBeginDocument{%
  \providecommand\BibTeX{{%
    \normalfont B\kern-0.5em{\scshape i\kern-0.25em b}\kern-0.8em\TeX}}}

\acmConference[AI4FashionSC KDD]{Virtual Event}{August 23, 2020}{Virtual Event, CA, USA}

\begin{document}

\title{Price Optimization in Fashion E-commerce}


\author{Sajan Kedia }
\email{sajan.kedia@myntra.com} 
\affiliation{%
  \institution{Myntra Designs}
  \country{India}
}
\author{Samyak Jain}
\email{jainsamyak2512@gmail.com}
\affiliation{%
 \institution{Myntra Designs}
 \country{India}}
 
\author{Abhishek Sharma}
\email{abhishek.sharma4@myntra.com}
\affiliation{%
 \institution{Myntra Designs}
 \country{India}}

\renewcommand{\shortauthors}{Kedia, et al.}

\begin{abstract}
With the rapid growth in the fashion e-commerce industry, it is becoming extremely challenging for the E-tailers to set an optimal price point for all the products on the platform. By establishing an optimal price point, they can maximize the platform's overall revenue and profit. In this paper, we propose a novel machine learning and optimization technique to find the optimal price point at an individual product level. It comprises three major components. Firstly, we use a demand prediction model to predict the next day\textquotesingle s demand for each product at a certain discount percentage. Next step, we use the concept of price elasticity of demand to get the multiple demand values by varying the discount percentage. Thus we obtain multiple price demand pairs for each product and we\textquotesingle ve to choose one of them for the live platform. Typically fashion e-commerce has millions of products, so there can be many permutations. Each permutation will assign a unique price point for all the products, which will sum up to a unique revenue number. 
To choose the best permutation which gives maximum revenue, a linear programming optimization technique is used. We\textquotesingle ve deployed the above methods in the live production environment and conducted several A/B tests. According to the A/B test result, our model is improving the revenue by 1\% and gross margin by 0.81\%.
\end{abstract}

\keywords{Pricing, Demand Prediction, Price Elasticity of Demand, Linear Programming, Optimization, Regression}

\maketitle

\section{Introduction}
The E-commerce industry is growing at a very rapid rate. According to Statista report\footnote{\url{https://www.statista.com/statistics/379046/worldwide-retail-e-commerce-sales/}}, it is expected to become \$4.88 trillion markets by 2021. One of the biggest challenges to the e-commerce companies is to set an optimal price for all the products on the platform daily, which can overall maximize the revenue \& the profitability. Specifically fashion e-commerce is very price sensitive and shopping behaviour heavily relies on discounting. Typically in a fashion e-commerce platform, starting from the product discovery to the final order, discount and the price point is the main conversion driving factor. As shown in Figure 1, the List Page \& product display page (PDP) displays the product(s) with it\textquotesingle s MRP and the discounted price. Product 1,4 is discounted, whereas product 2,3 is selling on MRP. This impacts the click-through rate (CTR) \& conversion. Hence, finding an optimal price point for all the products is a critical business need which maximizes the overall revenue.

To maximize revenue, we need to predict the quantity sold of all products at any given price. To get the quantity sold for all the products, a Demand Prediction Model has been used. The model is trained on all the products based on their historical sales \& browsing clickstream data. The output label is the quantity sold value, which is continuous. Hence it\textquotesingle s a regression problem. Demand prediction has a direct impact on the overall revenue. Small changes in it can vary the overall revenue number drastically. For demand prediction, several regression models like Linear Regression,  Random Forest Regressor, XGBoost \cite{chen2016xgboost}, MultiLayer Perceptron have been applied. An ensemble \cite{qiu2014ensemble} of the models mentioned above is used since each base model was not able to learn all aspects about the nature of demand of the product and wasn\textquotesingle t giving adequate results. Usually, the demand for a product is temporal, as it depends on historical sales. To leverage this fact, Time Series techniques like the autoregressive integrated moving model (ARIMA) \cite{li2018forecasting} \cite{contreras2003arima} is used. An RNN architecture based LSTM(Long Short-Term Memory) model \cite{greff2016lstm} is also used to capture the sequence of the sales to get the next day\textquotesingle s demand. In the result section, a very detailed comparative study of these models is explained.

Another significant challenge is cannibalization among products, i.e., if we decrease the discount of a product, it can lead to an increase in sales for other products competing with it. For example, if the discount is decreased for Nike shoes, it can lead to a rise in sales for Adidas or Puma shoes since they are competing brands \& the price affinity is also very similar. We overcame this problem by running the model at a category level and creating features at a brand level, which can take into account cannibalization. The second major challenge is to predict demand for new products, it\textquotesingle s a cold start problem since we don\textquotesingle t have any historical data for these products. To overcome this problem, we\textquotesingle ve used a Deep Learning-based model to learn the Product embedding. These embeddings are used as features in the Demand Model.

The demand prediction model generates demand for all the products for tomorrow at the base discount value. To get demand at different discount values economics concept of "Price Elasticity of Demand" (PED) is used. Price elasticity represents how the demand for a product varies as the price changes. After applying this, we get multiple price-demand pairs for each product. We have to select a single price point for every product such that the chosen configuration maximizes the overall revenue. 
To solve it, the Linear Programming optimization technique \cite{solow2007linear} is used. Linear Programming Problem formulation, along with the constraints, is described in detail in section 3.4.

The solution was deployed in a live production environment, and several A/B Tests were conducted. During the A/B test, A set was shown the same default price, whereas the B set was shown the optimal price suggested by the model. As explained in the Result section we\textquotesingle re getting a revenue improvement by approx 1\% and gross margin improvement by 0.81\%.

The rest of the paper is organized as follows. In Section 2, we briefly discuss the related work. We introduce the Methodology in Section 3 that comprises feature engineering, demand prediction model, price elasticity of demand, and linear programming optimization. In Section 4, Results and Analysis are discussed, and we conclude the paper in Section 5.

\begin{figure}[h]
\centering
\includegraphics[width=\linewidth]{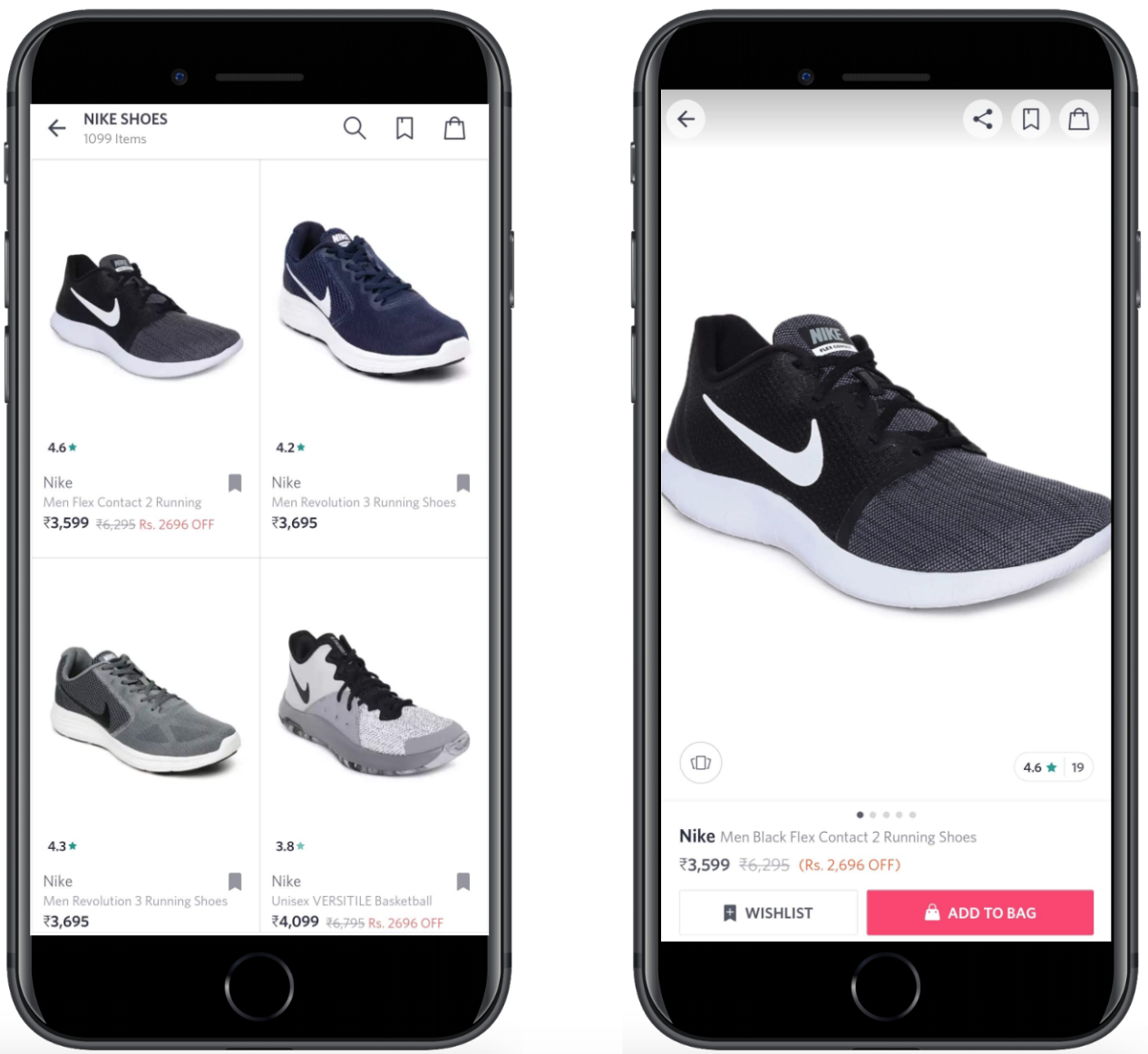}
\caption{List Page and PDP page}
\label{figure:action_items}
\end{figure}

\section{Related Work}
In this section, we\textquotesingle ve briefly reviewed the current literature work on Price Optimization in E-commerce. In \cite{Gupta2014AML}, authors are first creating customer segments using K-means clustering; then, a price range is assigned to each cluster using a regression model. Post that authors are trying to find out the customer\textquotesingle s likelihood of buying a product using the previous outcome of user cluster \& price range. The customer\textquotesingle s likelihood is calculated by using the logistic regression model. This model gives price range at a user cluster level, whereas in typical fashion e-commerce exact price point of each product is desired. 

Paper \cite{Singh2015DynamicPP} describes the price prediction methodology for AWS spot instances. To predict the price for next hour, they\textquotesingle ve trained a linear model by taking the last three months \& previous 24 hours of historical price points to capture the temporal effect. The weights of the linear model is learned by applying a Gradient descent based approach. This price prediction model works well in this particular case since the number of instances in AWS is very limited. In contrast, in fashion e-commerce, there can be millions of products, so the methodology mentioned above was not working well in our case.

In paper\cite{schlosser2018dynamic}, authors studied how the sales probability of the products are affected by customer behaviors and different pricing strategies by simulating a Test Market Environment. They used different learning techniques to estimate the sales probability from observable market data. In the second step, they used data-driven dynamic pricing strategies that also took care of the market competition. They compute the prices based on the current state as well as taking future states into account. They claim that their approach works even if the number of competitors is significant.

Paper \cite{ye2018customized} authors have explained the optimal price recommendation model for the hosts at the Airbnb platform. Firstly they\textquotesingle ve applied a binary classification model to predict booking probability of each listing in the platform; after that, a regression model predicts the optimal price for each listing for a night. At last, they apply a personalization layer to generate the final price suggestion to the hosts for their property. This setup is very different from a conventional fashion e-commerce setup where we need to decide an exact price point for all the products, whereas in the former case, it\textquotesingle s a range of price.

To address the above issues, we\textquotesingle ve proposed a novel machine learning mechanism to predict the exact price point for all the products on a fashion e-commerce platform at a daily level.

\section{Methodology}
In this section, we have explained the methodology of deriving the optimal price point for all the products in the fashion e-commerce platform.

\begin{figure}[h]
\centering
\includegraphics[width=\linewidth]{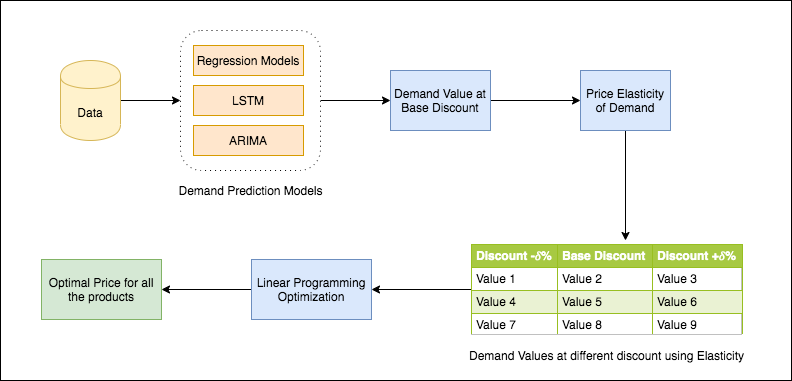}
\caption{Architecture Diagram}
\label{figure:action_items}
\end{figure}

Figure 2 shows the architecture diagram to optimize the price for all the products.

\begin{equation} \label{Revenue}
R = \sum_{i=1}^{n}p_{i}q_{i}
\end{equation}

Here R is the revenue of the whole platform, $p_{i}$ is the price assigned to $i^{th}$ product, and $q_{i}$ is the quantity sold at that price.

According to Equation 1, revenue is highly dependent on the quantity sold. Quantity sold is, in turn, dependent on the price of the product. Typically, as the price increases, the demand goes down and vice-versa. So there is a trade-off between cost and demand. To maximize revenue, we need demand for all the products at a given price. Therefore, a demand prediction model is required. Predicting demand for tomorrow at an individual product level is a very challenging task. For new products, there is a cold start problem which we address by using product embeddings. Another challenging task is how to capture the cannibalization of product.

\subsection{Feature Engineering}
To estimate the demand for all products in the platform extensive feature engineering is done. The features are broadly categorized into four categories :

\begin{figure}[h]
\centering
\includegraphics[width=\linewidth]{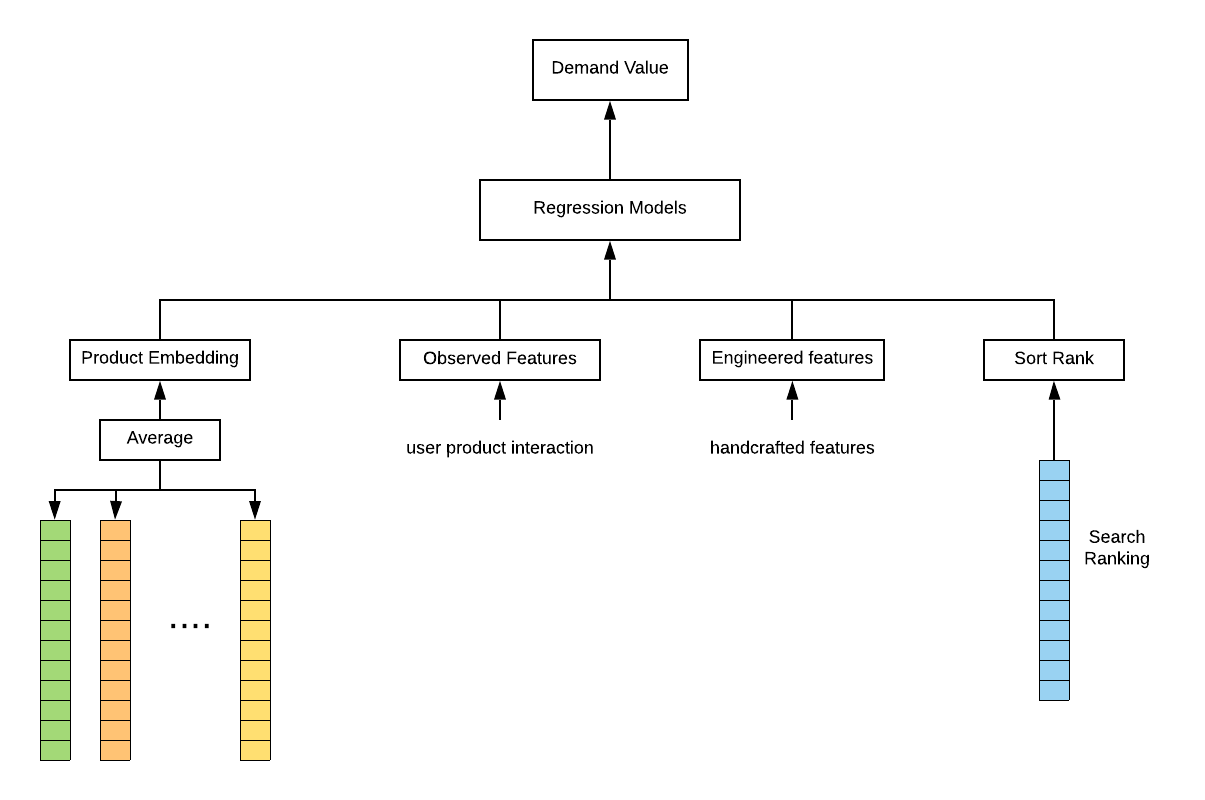}    
\caption{Feature Engineering for Demand Prediction}
\label{figure:action_items}
\end{figure}

\begin{itemize}
\item \textbf{Observed features}:  
These features showcase the browsing behavior of people. They include quantity sold, list count, product display count, cart count, total inventory, etc. From these features, the model can learn about the visibility aspects of the products. These directly correlate with the demand for the product.

\item \textbf{Engineered features}: 
To capture the historical trend of the products, we\textquotesingle ve handcrafted multiple features. For example, the last seven days of sales, previous 7 days visibility, etc. They also capture the seasonality effects and its correlation to the demand of the product. The ratio of quantity sold at BAG level (brand, article, gender) to the total quantity sold of a product is calculated to take care of the competitiveness among substitutable products.

\item \textbf{Sort rank}: Search ranking of a product is very highly correlated with the quantity sold. Products listed in the top search result generally have higher amounts sold. The search ranking score has been used for all the live products to capture this effect in our model.

\item \textbf{Product Embedding}: From the clickstream data, the user-product interaction matrix is generated. The value of an element in this matrix signifies the implicit score (click, order count, etc.) of a user\textquotesingle s interaction with the product. Skip-gram based model\cite{guthrie2006closer} is applied to the interaction matrix, which captures the hidden attributes of the products in the lower dimension latent space. Word2Vec \cite{rong2014word2vec} is used for the implementation purpose.
\end{itemize}

\subsection{Demand Prediction Model}
Demand for a product is a continuous variable, hence regression based models, sequence model LSTM, and time series model ARIMA were used to predict future demand for all the products. 
Since the data obeyed most of the assumptions of linear regression (the linear relationship between input and output variables,  feature independence, each feature following normal distribution and homoscedasticity), linear regression became an optimal choice. Hyperparameters like learning rate (alpha) and max no. of iterations were tuned using GridSearchCV and by applying 5 fold cross-validation. The learning rate of 0.01 and the max iteration of 1000 was chosen as a result of it. Overfitting was taken care of by using a combination of L1 and L2 regularization (Elastic Net Regularization). Tree-based models were also used. They do not need to establish a functional relationship between input features and output labels, so they are useful when it comes to demand prediction. Also, trees are much better to interpret and can be used to explain the outcome to business users adequately. Various types of tree-based models like  Random Forest and XGBoost were used. Hyperparameters like depth of the tree and the number of trees were tuned using the technique mentioned above. Overfitting was taken care of by post-pruning the trees.
Lastly, a MLP regressor with a single hidden layer of 100 neurons, relu activation, and Adam optimizer was used. Finally, an ensemble of all the regressors was adopted. The median of the output of all the regressors was chosen as the final prediction. It was chosen because each of the individual regressors was not able to learn all aspects of the nature of the demand for the product and wasn\textquotesingle t giving adequate results. Thus, the ensemble approach was able to capture all aspects of the data and delivered positive results.
LSTM with a hidden layer was also used to capture the temporal effects of the sale. Batch-size for each epoch was chosen in such a way that it comprised of all the items of a particular day. Hyperparameters like loss function, number of neurons, type of optimization algorithm, and the number of epochs were tuned by observing the training and validation loss convergence. Mean-squared error loss, 50 neurons, Adam optimizer, and 1000 epochs were used to perform the task.
Arima was also used to capture the trends and seasonality of the demand for the product.  The three parameters of ARIMA : p (number of lags of quantity sold to be used as predictors), q (lagged forecast errors), and d  (minimum amount of difference to make series stationery) were set using the ACF and PACF plots. The order of difference (d) = 2 was chosen by observing when the ACF plot approaches 0. The number of lags (p) = 1 was selected by observing the PACF plot and when it exceeds the PACF\textquotesingle s plot significance limit. Lagged forecast errors (q) = 1 was chosen the same way as p was chosen but by looking at the ACF plot instead of PACF.

\subsection{Price Elasticity of Demand}
From the demand model, we get the demand for a particular product for the next day based on its base discount. To get the demand for a product at a different price point, we use the concept of price elasticity of demand \cite{babar2015development}\cite{minga2003dynamic}.

\begin{figure}[h]
\centering
\includegraphics[width=\linewidth]{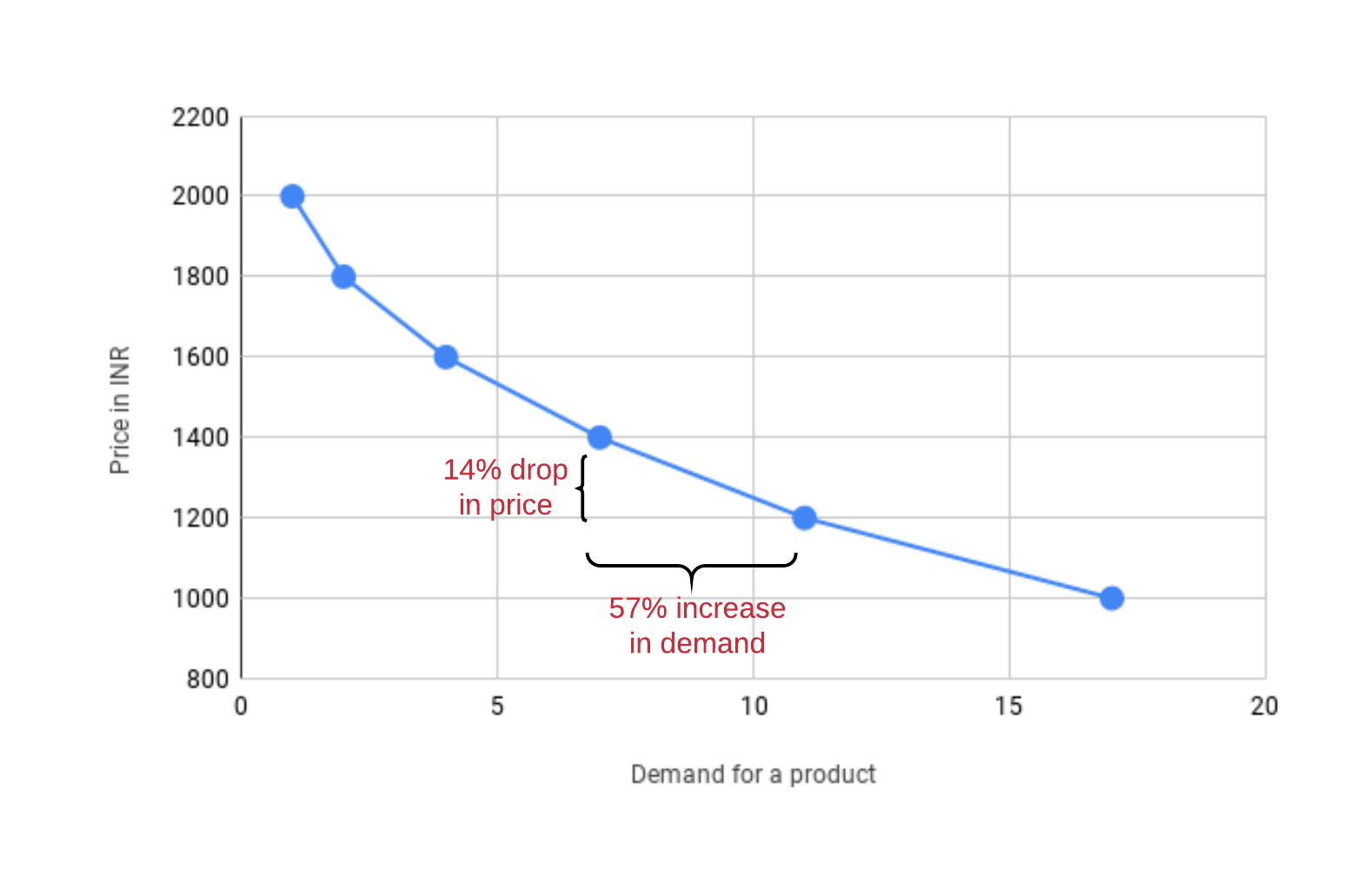}
\caption{Variation of Price and Demand of a Product}
\label{figure:action_items}
\end{figure}

Figure 4 illustrates the effects of change in price on the quantity demanded of a product. When the price drops from INR 1400 to 1200, the quantity demanded of the product increases from 7 to 11. The price elasticity of demand is used to capture this phenomenon.

The formula for price elasticity of demand is :


\begin{equation}\label{elasticity}
Ed = {\dfrac{\Delta{Q}} { Q}} . {\dfrac{P}{\Delta{P} }}
\end{equation}

where $Q$, $P$ are the original demand and price for the product, respectively. $ \Delta{Q}$ represents the change in demand as the price changes, and $\Delta{P}$ depicts the change in the product's price. 

A product is said to be highly elastic if for a small change in price there is a huge change in demand, unitary elastic if the change in price is directly proportional to change in demand and is called inelastic if there is a significant change in price but not much change in demand.

Since we are operating in a fashion e-commerce environment determination of price elasticity of demand is a challenging task. There is a wide range of products that are sold, so we experience all types of elasticity. Usually price and demand share an inverse relationship i.e., when the price decreases, the demand for that product increases and vice-versa.

But this relationship is not followed in the case of Giffen/Veblen goods. Giffen/Veblen goods are a type of product, for which the demand increases when there is an increase in the price. In our environment, we not only experience different types of elasticity, but we also have to deal with Giffen/Veblen goods. E.g., Rolex watch is a type of Veblen good as its demand increases with an increase in price because it is used more as a status symbol.

Due to these reasons calculating the price elasticity of demand at a brand, category, or any other global level is not possible. So we have computed it for each product based on the historical price-demand pairs for that product. For new products, the elasticity of similar products determined by product embeddings were used.

Price elasticity is used to calculate the product demand at different price points. From the business side, there is a strict guardrail of max $\pm \delta$ \% change in the base discount, where $\delta$ is a user defined threshold.  
The value of $\delta$ is kept as a small positive integer since a drastic change in the discounts of the products is not desired. We have the demand value for all the products for tomorrow at the base discount by using the aforementioned demand prediction model. We also calculate two more demand values by increasing the base discount by $\delta$\% and decreasing by $\delta$\% by using the price elasticity. Thus the output after applying the elasticity concept is that we get three different price points and the corresponding demand value for all products. The price elasticity changes with time, so its value is updated daily.

\begin{figure*}
   \includegraphics[width=\textwidth,scale=2]{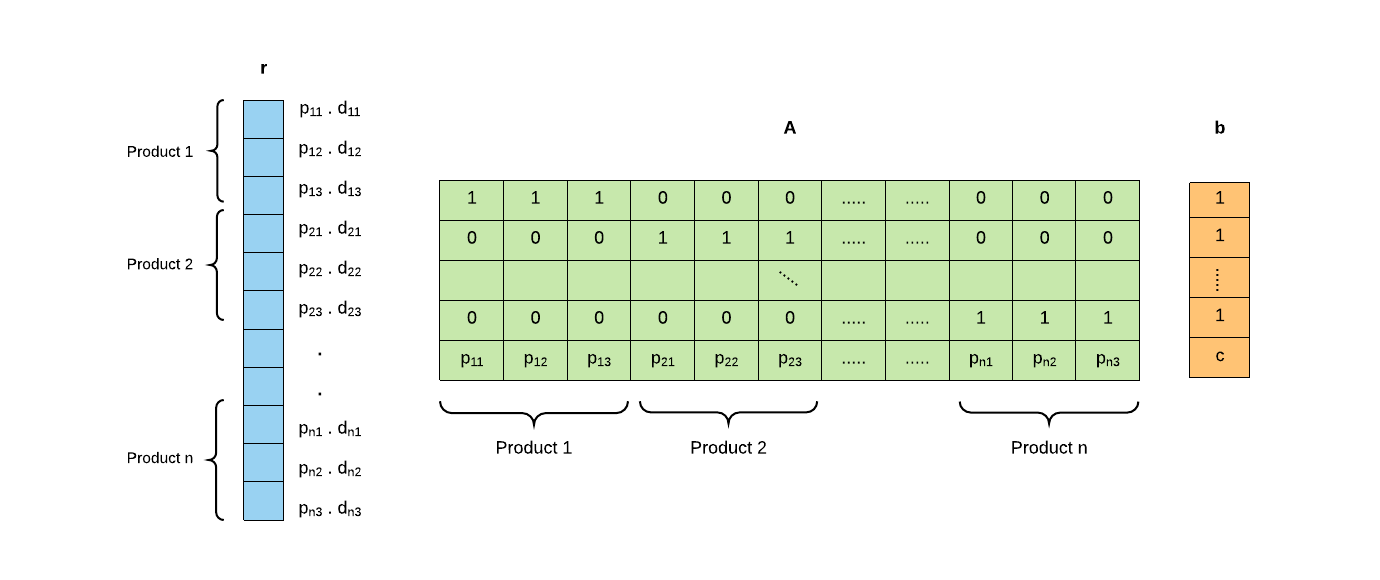}
   \captionsetup{justification=centering}
   \caption{Linear Programming Vector Computations}
   \label{figure:1}
\end{figure*}

\subsection{Linear Programming}
After applying the price elasticity of demand, we get three different price points for a product and the corresponding demand at those price points.

Now we need to choose one of these three prices such that the net revenue is maximized. Given $3$ different price points and $N$ products, there will be $3^N$ permutations in total. Thus, the problem boils down to an optimization problem, and we formulate the following integer problem formulation:
\cite{gomory1960integer}. 

\[\textbf{max}\sum_{k=1}^{3}\sum_{i=1}^{n} p_{ik}x_{ik}d_{ik}\]\\
 \centerline{\textbf{Subject To the constraints}} 
\[\sum_{k=1}^{3} x_{ik} = 1 \; \forall \; i \in n\]
\[\sum_{k=1}^{3}\sum_{i=1}^{n} p_{ik}x_{ik} = c\]
\[x_{ik} \in \{0,1\} \]\\
In the above equations $p_{ik}$ denotes the price and $d_{ik}$ denotes the demand for $i^{th}$ product if $k^{th}$ price-demand pair is chosen. Here, $x_{ik}$ acts like a mask that is used to select a particular price for a product. $x_{ik}$ is a boolean variable (0 or 1) where 1 denotes that we select the price and 0 that we do not select it. The constraint $\sum_{k=1}^{3} x_{ik}$ = 1 makes sure that we select only 1 price for a product and $\sum_{k=1}^{3}\sum_{i=1}^{n} p_{ik}x_{ik}$ = c makes sure that sum of all the selected prices is equal to c, where c is user-defined variable. C can be varied from its minimum value that is obtained when the least price for each product is chosen to its maximum value that is obtained when the highest price for each product is chosen. Somewhere in between, we get the optimal solution.

The above formulation is computationally intractable since there can be millions of permutations of price-demand pairs. Thus we convert the above integer programming problem to a linear programming problem by treating $x_{ik}$ as a continuous variable. This technique is called linear relaxation. The solution obtained from this problem acts like a linear bound for the integer formulation problem. Here, for the $i^{th}$ product the price corresponding to maximum value of $x_{ik}$ is chosen as the optimal price point. The linear programming problem \cite{dempster1999pricing} thus obtained is :


\[\textbf{max} \sum_{k=1}^{3}\sum_{i=1}^{n} p_{ik}x_{ik}d_{ik}\]\\

\centerline{\textbf{Subject To the constraints}}

\[\sum_{k=1}^{3} x_{ik} = 1 \; \forall \; i \in n\]

\[\sum_{k=1}^{3}\sum_{i=1}^{n} p_{ik}x_{ik} = c\]

\[ 0 \leq x_{ik} \leq 1  \]\\

We implement the above linear programming problem using Scipy linear programming library routine linprog. The input is given in the vector form as follows:\\


\centerline{max \textbf{r.x}}

\centerline{Subject To \textbf{A.x = b}}

Where $r$ is a vector that contains the revenue contributed by a product at each price point, $ A$ is a matrix that contains information about the products and their prices.

As shown in Figure 5, a set of 3 columns represents a product. In this set of 3 columns, the first column represents value corresponding to (discount-$\delta$\%), middle column to base discount, last column to (discount+$\delta$\%). The last row of $A$ contains different price points according to their updated discount value \& product. $b$ vector contains the upper bound of the constraints. All entries in $b$ are 1 except the last one, which is equal to c. Here 1 represents that only one price can be selected for each product, and c indicates the summation corresponding to the chosen prices.

\section{Results \& Analysis}

In this section, we have discussed the data sources used to collect the data, followed by an analysis of the data and the insights derived from it. Different kinds of evaluation metrics used to check the accuracy are also discussed along with the comparative study of different models used.

\subsection{Data Sources}
Data was collected from the following sources : 
\begin{itemize}

\item \textbf{Clickstream data}: this contained all user activity such as clicks, carts, orders, etc.
\item \textbf{Product Catalog}: this contained details of a product like brand, color, price, and other attributes related to the product.
\item \textbf{Price data}: this contained the price and the quantity sold of a product at hour level granularity.
\item \textbf{Sort Rank}: this contained search rank and the corresponding scores for all the live products on the platform.

\end{itemize}

\subsection{Analysis of Data}
According to the analysis, 20\% of the products generate 80\% of the revenue. Hence, it is required to price these products precisely. This is the biggest challenge when it comes to demand prediction.
\begin{figure}[h]
\centering
\includegraphics[width=\linewidth]{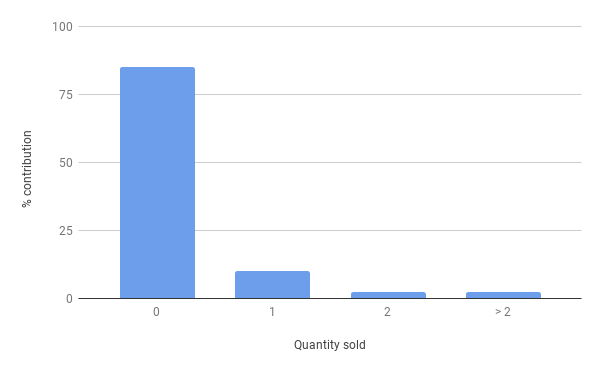}
\caption{Distribution of Quantity Sold}
\label{figure:action_items}
\end{figure}
Figure 6 shows the distribution of quantity sold for all the products on the platform. It is evident that a minority of products contribute towards the total revenue.

Figure 7 shows the elasticity distribution of the products on the platform. The y-axis of the graph depicts the density rather than the actual count frequency of the elasticity value. As discussed in Section 3.3, the determination of elasticity is a difficult task. All kinds of elasticity are experienced, and hence elasticity is calculated at a product level. It is evident from the figure that the range of elasticity is from -5 to +5. -5 indicates that if the price is dropped, then demand increases drastically, and +5 suggests that if the price is increased, then demand also increases by a large margin. It can be observed that most of the time, elasticity value is 0; this is so because 80\% of the quantity of the product sold is 0. Lastly, it can be concluded that most of the product's elasticity lies between -1 to +1. Hence, most of the products are relatively inelastic, but there are few products whose elasticity is very high and are at extreme ends, so it has to be calculated accurately. 

\begin{figure}[h]
\centering
\includegraphics[width=\linewidth]{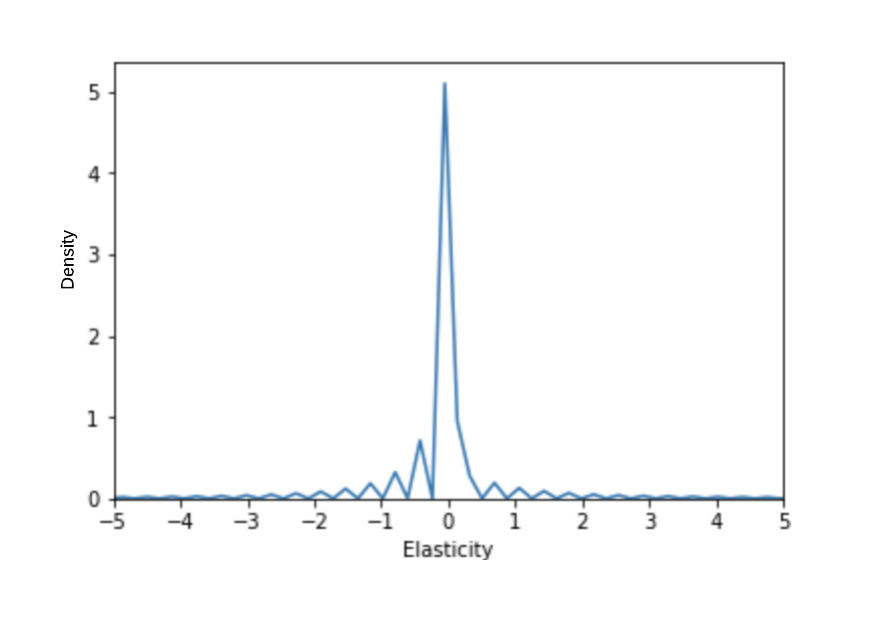}
\caption{Distribution of Elasticity}
\label{figure:action_items}
\end{figure}

\subsection{Evaluation Metric}
To evaluate the demand prediction models, mean absolute error \& root mean squared error is used as a metric. They both tell us the idea of how much the predicted output is deviating from the actual label. For this scenario, coefficient of determination i.e., $R^{2}$ or adjusted $R^{2}$, is not a good measure as it involves the mean of the actual label. Since 80\% of the time actual label is 0 mean cannot accommodate it, and hence $R^{2}$ cannot judge the demand model\textquotesingle s performance. 
  
\subsection{Results}

In this section results of different models are discussed in table 1. Majorly three different classes of the model were tested and compared. They include regressors, LSTM, and ARIMA. \textit{mae} and \textit{rmse} were chosen as the evaluation metric to compare these models. The reason for this choice is discussed in section 4.3.
All the models were trained on historical data of the past three months, comprising of over a million records. For the test data, the recent single day records were used. First, different kinds of regressors were used; out of all the regressors, XGBoost gave the best result. It is so because it's tough to establish a functional relationship between input features and output demand. As XGBoost does not need to establish any functional relationship, it performed better. However, all individual regressors were not able to capture all aspects of the data, so an ensemble of all the regressors gave the ideal result. The ensemble takes advantage of all regressors and hence is chosen. Then, after regressors, LSTM \& ARIMA were also tried, but it failed to give the desired result as due to multiple external factors in the business, the demand data for the products did not exhibit sequential \& temporal characteristics.

\begin{table}[]
\begin{tabular}{|l|l|l|}
\hline
\textbf{Model}                  & \textbf{mae} & \textbf{rmse} \\ \hline
Linear Regression               & 0.207        & 0.732         \\ \hline
Random Forest Regressor         & 0.219        & 0.854         \\ \hline
XG Boost                        & 0.195        & 0.847         \\ \hline
MLP Regressor & 0.254        & 1.471         \\ \hline
Ensemble                        & 0.192        & 0.774         \\ \hline
LSTM                            & 0.221        & 0.912         \\ \hline
ARIMA                           & 0.258        & 1.497         \\ \hline
\end{tabular}
\caption{\label{table:1}\textit{Comparative performance of various models}}
\end{table}

    

\subsection{Experimental Design}
In this section, we describe live experiments that were performed on one of the largest fashion e-commerce platforms.
The experiments were run on around two hundred thousand styles spread across five days.
To test the hypothesis that model recommended prices are better than baseline prices, two user groups were created:\\
\begin{itemize}
\item Set-A (Control group) was shown the baseline prices 
\item Set-B (Treatment group) was shown model recommended prices.\\
\end{itemize}

These sets were created using a random assignment from the set of live users. 50\% of the total live users were assigned to Set A and the rest to Set B.

Users in Control and Treatment groups were exposed to the same product. But the products were priced differently according to the group they belong to. Both groups were compared with respect to the overall platform revenue and gross margin. Revenue is already defined and explained in section 3, whereas gross margin can be defined as follows :

\[Gross Margin = (Revenue - (buying\; cost)) / Revenue\] 

In general, the gross margin in our scenario is pure profit bottom line.

\begin{table}[h!]
\begin{center}
\begin{tabular}{|c| c | c|} 
\hline
 & \textit{Percentage increment in Revenue} & \textit{GM \% Uplift} \\ 
\hline
\textit{Test 1} & 0.96\% & 0.99\% \\ 
\textit{Test 2} & 1.96\% & 0.95\% \\
\textit{Test 3} & 0.09\% & 0.49\% \\
\textit{Test 4} & 3.27\% & -0.41\% \\
\textit{Test 5} & 7.05\% & 0.15\% \\
\hline
\end{tabular}
\caption{\label{table:2}\textit{A/B test results}}
\end{center}
\end{table}

In table 2, the first three instances correspond to the business unit - Men\textquotesingle s Jeans and Streetwear, while the last 2 are of  Women\textquotesingle s Western wear and Eyewear respectively. 

In the case of Men\textquotesingle s Jeans and Streetwear, there is a steady increase in both revenue and gross margin of the whole platform. From this, it can be concluded that the model recommended price gave positive results for this business unit.
In the case of Women\textquotesingle s Western wear, there was a massive lift in revenue, but the gross margin fell. The reason for this is that women\textquotesingle s products were highly elastic. Small changes in price had a significant impact on demand. So due to the model's recommended price, there was a vast fluctuation in demand (i.e., demand for products increased), and the revenue thus increased. However, the gross margin was slightly impacted, and it decreased.
 
Eyewear is a small business unit, so the impact of discount change is enormous, i.e., 7.05\% increment in the revenue, whereas the GM increased by 0.15\%.

Overall there was an approximately 1\% increase in revenue of the platform and 0.81\% uplift in gross margin due to model recommended prices. These are computed by taking the average of the first three instances since they belong to the same business unit.

\section{Conclusion}
Fashion e-tailers currently find it extremely hard to decide an optimal price \& discounting for all the products on a daily basis, which can maximize the overall net revenue \& the profitability. To solve this, we have proposed a novel method comprising three major components. First, a demand prediction model is used to get an accurate estimate of tomorrow\textquotesingle s demand. Then, the concept of price elasticity of demand is used to get the demand for a product at multiple price points. Finally, a linear programming optimization technique is used to select one price point for each product, which maximizes the overall revenue. Online A/B test experiments show that by deploying the model revenue and gross margin increases. Right now, the prices are decided and updated daily, but in the future, we would like to extend our work so that we can also capture the intra-day signals in our model \& accordingly do the intra-day dynamic pricing.

\bibliographystyle{ACM-Reference-Format}
\bibliography{sample-base}

\appendix

\end{document}